\documentclass[10pt,twocolumn,letterpaper]{article}

\usepackage{wacv}              

\usepackage{graphicx}
\usepackage{amsmath}
\usepackage{amssymb}
\usepackage{booktabs}
\usepackage{multirow}
\usepackage{placeins}
\usepackage[accsupp]{axessibility}
\usepackage[pagebackref,breaklinks,colorlinks]{hyperref}

\usepackage[capitalize]{cleveref}
\crefname{section}{Sec.}{Secs.}
\Crefname{section}{Section}{Sections}
\Crefname{table}{Table}{Tables}
\crefname{table}{Tab.}{Tabs.}

\def\wacvPaperID{130} 
\def\confName{WACV}
\def\confYear{2025}

\begin{document}

\title{VIIS: Visible and Infrared Information\\ Synthesis for Severe Low-light Image Enhancement}




\author{
    Chen Zhao$^{1}$\thanks{Equal Contribution.} \quad
    Mengyuan Yu$^{2}$\footnotemark[1] \quad
    Fan Yang$^{1}$ \quad
    Peiguang Jing$^{1}$\thanks{Corresponding author} \\[1ex]
    $^1$Tianjin University, Tianjin, China \\
    $^2$Southeast University, Nanjing, China \\[1ex]
    \tt\small\texttt{\{z\_c, eeyf, pgjing\}@tju.edu.cn} \quad \tt\small\texttt{yumengyuan@seu.edu.cn} \\
}

\maketitle

\begin{abstract}
Images captured in severe low-light circumstances often suffer from significant information absence. Existing singular modality image enhancement methods struggle to restore image regions lacking valid information. By leveraging light-impervious infrared images, visible and infrared image fusion methods have the potential to reveal information hidden in darkness. However, they primarily emphasize inter-modal complementation but neglect intra-modal enhancement, limiting the perceptual quality of output images. To address these limitations, we propose a novel task, dubbed visible and infrared information synthesis (VIIS), which aims to achieve both information enhancement and fusion of the two modalities. Given the difficulty in obtaining ground truth in the VIIS task, we design an information synthesis pretext task (ISPT) based on image augmentation. We employ a diffusion model as the framework and design a sparse attention-based dual-modalities residual (SADMR) conditioning mechanism to enhance information interaction between the two modalities. This mechanism enables features with prior knowledge from both modalities to adaptively and iteratively attend to each modality's information during the denoising process. Our extensive experiments demonstrate that our model qualitatively and quantitatively outperforms not only the state-of-the-art methods in relevant fields but also the newly designed baselines capable of both information enhancement and fusion. The code is available at \href{https://github.com/Chenz418/VIIS}{https://github.com/Chenz418/VIIS}.
\end{abstract}

\section{Introduction}
\label{sec:intro}
Low-light image enhancement aims to ameliorate the perceptual quality and visibility of images captured under low-light conditions. This task is important not only for human visual perception but also for downstream vision tasks such as object detection \cite{liu2023image} and automatic driving \cite{li2021deep}. With the development of deep learning, numerous learning-based methods \cite{a2024learning,jin2022unsupervised,Cai_2023_ICCV,zamir2020learning,zhou2022lednet,sharma2021nighttime,jiang2021enlightengan,hou2024global} have demonstrated promising performance in addressing fundamental challenges in low-light images (\eg, low brightness, low contrast, and other basic corruptions).

However, in certain realistic scenarios characterized by exceptionally low ambient light (\eg, agricultural monitoring, subterranean exploration, and wildlife photography), captured images are often dominated by heavy shadows and noise, obscuring abundant information. Under these conditions, low-light image enhancement methods based solely on visible images become highly ill-posed, as some severely affected regions may lack essential original information. Consequently, these methods demonstrate significant limitations in their effectiveness. 

\begin{figure}
\centering
\includegraphics[width=0.9\linewidth]{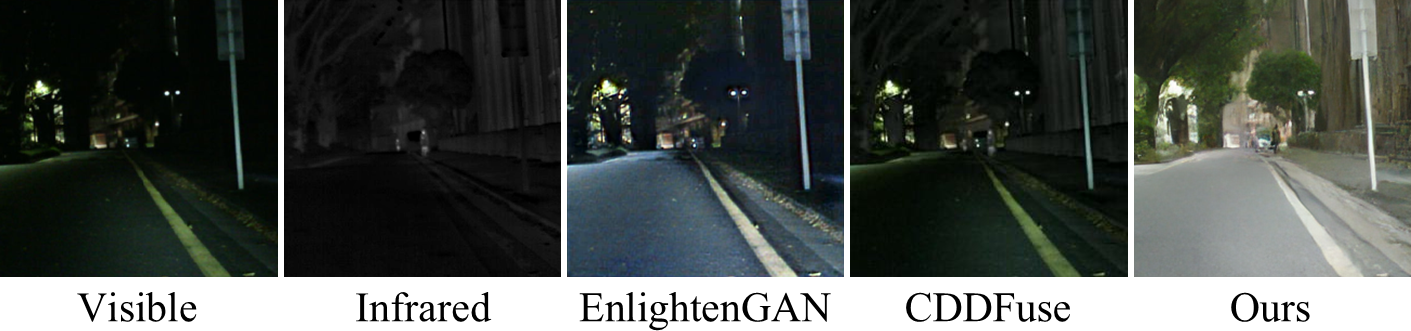}
   \caption{An existing low-light enhancement method EnlightenGAN \cite{jiang2021enlightengan} fails to effectively enhance the image background where most information is obliterated by darkness. The visible and infrared image fusion method CDDFuse \cite{Zhao_2023_CVPR} reveals the outlines of people and buildings, however, the overall image remains low-luminosity and the region complemented by the infrared image remains sketchy.}
\label{fig:Instance}
\end{figure}

Resorting to infrared images, which remain impervious to low light, is a promising way to reveal information corrupted by darkness. Infrared imaging has already been widely used in practical low-light situations such as certain surveillance systems \cite{ibrahim2016comprehensive} and wildlife photography \cite{cilulko2013infrared, burke2019optimizing}. However, the inherent properties of infrared images limit their ability to convey color and detailed texture information. The visible and infrared image fusion (VIF) methods \cite{Zhao_2023_ICCV,huang2022reconet,liang2022fusion,wang2024cs2fusion,li2024crossfuse} merge merits from both infrared and visible modalities, but they also exhibit non-negligible limitations in low-light scenarios. Specifically, given that VIF methods are designed to merge complementary information or best parts from input images \cite{liang2022fusion}, most of them merely focus on inter-modality information complementation but neglect intra-modality information enhancement. Hence, in low-light RGB circumstances, the visible image information in the fused images tends to remain dark and low-contrast, mirroring its original representation. Additionally, the infrared images fail to utilize the color and texture information from visible images, resulting in grayscale and sketchy representation in the fused images. These drawbacks significantly affect the perceptual quality of the fused images. Although DIVFusion \cite{tang2023divfusion} jointly enhances visible images and facilitates complementary
information aggregation. However, it neglects the enhancement for the infrared modality and, for the visible modality, it only enhances the luminance (Y) channel. \cref{fig:Instance} illustrates the limitations of singular modality image enhancement and VIF methods in severely low-light situations.

To explicitly unveil situation information in severely low-light circumstances and enhance the original information to achieve outputs with superior perceptual quality, we propose a novel task, visible and infrared information synthesis (VIIS). The task is designed to achieve both dual-modal information enhancement and fusion. Due to the difficulty of obtaining ground truth in VIIS, we design an information synthesis pretext task (ISPT), which guides the model to enhance the overall visibility of visible images while concurrently complementing the obliterated regions with colorized infrared information.

Recently, diffusion models (DMs) \cite{sohl2015deep,ho2020denoising} have shown high performance in various fields. To leverage the advantageous properties of DMs, we utilize the diffusion model as the framework of the model. Since the interaction between the two modalities is crucial for VIIS, we design a sparse attention-based dual modalities residential (SADMR) conditioning mechanism to realize efficient multi-modal information interaction in the denoising process. Specifically, as many DM-based image editing models \cite{shen2023difftalk, joseph2024iterative} the conditioning images are initially concatenated with the original noise. Then, in each block of Unet-based denoising network, the encoded visible and infrared image features are parallelly injected into the embedded intermediate features. In addition, for better comprehension of the injected image features, we utilize a sparse cross-attention module to replace the classic adding or concatenating operation \cite{gao2023implicit, zhang2023adding}. Hence, in the denoising process, the intermediate features encapsulating prior knowledge from both modalities attend to each modality's features adaptively and iteratively, which enables sufficient multi-modal information interaction. 

The main contributions of this work are as follows:
\begin{itemize}
\setlength{\leftmargin}{0pt}
\item
We propose a novel task VIIS to enhance images in severely low-light circumstances. Compared to existing image enhancement and image fusion methods, it can unveil situation information hidden in darkness by resorting to infrared images while effectively enhancing original information to achieve high perceptual quality.

\item
We propose a SADMR conditioning mechanism. It enables the intermediate features of Unet-based denoising network, which encapsulates prior knowledge from both modalities, to adaptively attend to the features of each modality. This mechanism realizes effective multi-modal information interaction.

\item
We propose a pretext task to guide the model in achieving both inter-modal complementation and intra-modal enhancement. We conduct extensive quantitative and qualitative experiments on two public datasets MSRS \cite{tang2022piafusion} and KAIST-MS \cite{hwang2015multispectral}. The results demonstrate that our model can produce images possessing high perceptual quality while obtaining complementary information from source images. 

\end{itemize}

\section{Related works}

\subsection{Low-light image enhancement}
In the past few years, many learning-based methods have been developed to enhance low-light images \cite{zamir2020learning,sharma2021nighttime,jiang2021enlightengan,fu2022gan}. Recently, several investigations have managed to address varied specific challenges in this field. For instance, Jin \etal \cite{jin2022unsupervised} boost the intensity of dark regions while suppressing the light effects in bright regions with a layer decomposition network and a light-effects suppression network. Cai \etal \cite{Cai_2023_ICCV} employ Illumination-Guided Transformer (IGT) to enhance the low-light regions while restoring the corruptions like noise, artifacts, under-/over-exposure, and color distortion. Sharif \etal \cite{a2024learning} incorporate a lightweight and mobile-friendly deep network along with an optimization strategy to tackle the limitations for accelerating edge vision tasks. However, existing low-light image enhancement methods struggle to restore images in severely low-light situations and are unable to effectively address the challenge of substantial information loss in certain image regions.

\subsection{Visible and infrared image fusion}
VIF is a subset of multi-modality image fusion (MMIF). It leverages the immunity of infrared images to low-light circumstances, which has the potential to address the information absence in visible images. Deep learning-based VIF methods are categorized into three main groups: CNN-based methods\cite{li2021different, long2021rxdnfuse, zhang2020rethinking}, AE-based methods \cite{li2018densefuse, li2021rfn, xu2021classification}, and GAN-based methods \cite{MA201911, fu2021image, ma2020ddcgan}. Recently, Liang \etal \cite{liang2022fusion} propose a powerful image decomposition model for the fusion task via self-supervised representation learning.  Zhao \etal \cite{Zhao_2023_ICCV} propose to effectively extract cross-modality features by decomposing desirable modality-specific and modality-shared features. Wang \etal \cite{wang2024cs2fusion} considers infrared images as a complement to visible images and design a compensation perception network to generate fusion images by estimating the feature compensation map of infrared images. However, most VIF methods only emphasize inter-modality complementary but neglect intra-modality enhancement.

\subsection{Diffusion models}
Diffusion models (DMs) \cite{sohl2015deep,ho2020denoising} have demonstrated excellent efficacy as image-generative models. Numerous studies have managed to achieve improvements in various aspects, encompassing sampling guidance \cite{dhariwal2021diffusion,ho2022classifier}, iterations \cite{salimans2021progressive,song2020denoising,lu2022dpm}, and model computational complexity \cite{rombach2022high,ho2022cascaded}. To accommodate the variety of conditioning modalities, a flexible set of conditioning mechanisms has been developed. The simplest method is concatenation \cite{saharia2022palette}, where the condition is directly concatenated with the noisy feature. Rombach \etal \cite{rombach2022high} propose a conditioning mechanism that injects conditioning signals through cross-attention, enabling effective processing of sequential features. Hu \etal \cite{hu2023animate} propose a conditioning mechanism to preserve the consistency of intricate appearance features, which utilizes ReferenceNet to merge detailed features via spatial attention \cite{vaswani2017attention}. Zhang \etal \cite{zhang2023adding} and Li \etal \cite{li2024controlnet++} reuse the pre-trained encoding layers of the Unet-based denoising network to embed spatial conditionings. Processed by zero convolution layers, the embedded features are then added to the decoding layers of the denoising network.

Recently, several studies have extended DMs into image enhancement and image fusion. For low-light image enhancement, Zhou \etal \cite{ijcai2023p199} propose a novel pyramid diffusion method for low-light image enhancement, which performs sampling in a pyramid resolution style. Jiang \etal \cite{jiang2023low} present a wavelet-based conditional diffusion model, leveraging wavelet transformation to accelerate the inference process for low-light image enhancement. Hou \etal \cite{hou2024global} formulate a curvature regularization term anchored in the intrinsic non-local structures of image data in their diffusion model, which facilitates the preservation of complicated details and the augmentation of contrast for low-light image enhancement. For image fusion tasks, Zhao \etal \cite{Zhao_2023_ICCV} pioneers the application of DMs to tackle the problem of multi-modality image fusion, which integrates expectation-maximization (EM) into the diffusion sampling iteration. In this paper, we employ a diffusion model as the framework and propose a conditioning mechanism to handle multi-modal image inputs.

\section{Methodology}
\begin{figure*}[h]
\centering
\includegraphics[width=0.95\linewidth]{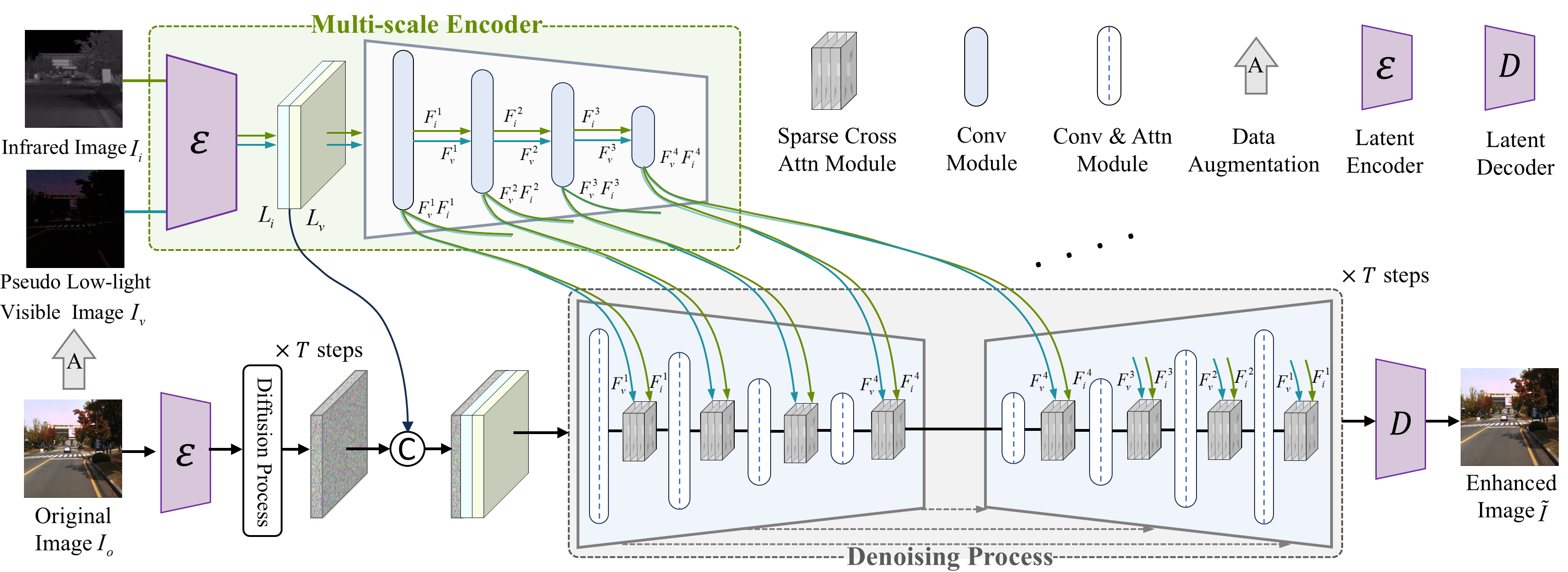}
   \caption{The overall of our model. The information synthesis pretext task (ISPT) initially generates the pseudo-low-light visible images with a data augmentation strategy. Subsequently, in one branch, the latent space infrared and visible images are concatenated with noise. In the other branch, these latent space images are encoded to obtain multi-scale features, which are then injected into the Unet-based denoising network through sparse cross-attention.}
\label{fig:overall}
\end{figure*}

\subsection{Overview}
Given a pair of visible and infrared images, our model is designed to enhance the visible images and complement the severely degraded regions by leveraging the colorized infrared images. During the process, the enhancement of visible images and colorization of infrared images are based on inter-modal information interaction and coordination. Our model utilizes the latent diffusion model (LDM) \cite{rombach2022high} as the framework, which is discussed in \cref{sec:Preliminary}. To achieve effective multi-modal information interaction, we propose SADMR conditioning mechanism in \cref{sec:conditioning}. Due to the difficulty in obtaining ground truth for VIIS task , we design an information synthesis pretext task to guide the training of the model, as detailed in \cref{sec:pretext}. \cref{fig:overall} shows the overall of our model.

\subsection{Preliminary}
\label{sec:Preliminary}
LDM \cite{rombach2022high} transforms input images into a spatially lower-dimensional latent space to reduce computational complexity. As many diffusion models (DMs) \cite{sohl2015deep,ho2020denoising}, it is trained to iteratively denoise a normally distributed variable by reversing a diffusion process. Specifically, in LDM, given an input image $ I \in \mathbb{R}^{H \times W \times 3} $, it is first encoded by a well-trained encoder \cite{esser2021taming} into a latent space $ {z_0} = \varepsilon (I) \in \mathbb{R}^{h \times w \times 3} $, where the encoder downsamples the image by a factor $ f = H/h = W/w $. During training, LDM defines a diffusion process in which Gaussian noise is iteratively introduced to the original features $ {z_0} $ according to the following Markov \cite{geyer1992practical} process:
\begin{equation}
	\begin{split}
         q\left( {{z_t}\mid {z_{t - 1}}} \right) = {\cal N}\left( {{z_t};\sqrt {{\alpha _t}} {z_{t - 1}},\left( {1 - {\alpha _t}} \right){\bf{I}}} \right),\forall t \in \{
         \\1,...,T\},
	\end{split}
 \label{eq:Markovian}
\end{equation}
where $ T $ is the number of diffusion steps, $ {z_t} $ is the noised version of $ {z_0} $, hyperparameter $ {\alpha _t} \in (0,1) $  representing the variance schedule across diffusion steps and $ {\bf{I}} $ is the identity matrix with the same dimensions as $ {z_0} $.

In the denoising process, the restoration of the original features $ {z_0} $ is achieved by reversing the diffusion process. On this basis, LDM is trained to reverse the diffusion process by minimizing the objective:
\begin{equation}
  L_{L D M}=\mathbb{E}_{z, \epsilon \sim \mathcal{N}(0, \mathbf{I}), C, t}\left[\left\|\epsilon-\epsilon_{\theta}\left(z_{t}, t, C\right)\right\|_{2}^{2}\right], 
  \label{eq:objective}
\end{equation}
where  $ C $ represents the collection of conditioning information, $ \epsilon_{\theta} $ is the network responsible for predicting the noise. The final denoised result $ {\widetilde z_0} $ is then upsampled to the pixel space with the pre-trained decoder $ \widetilde I = D({\widetilde z_0})$.

\subsection{SADMR conditioning mechanism}
\label{sec:conditioning}
The VIIS task necessitates the enhancement of visible information and the colorization of infrared information. However, independently enhancing and colorizing these two modalities renders the VIIS task an ill-posed problem. This is because of the one-to-many mapping involved in both enhancing severely corrupted visible images and colorizing infrared images. Hence, interaction and coordination between visible and infrared images is crucial, as it allows each modality to reference related information from the other. 

To achieve effective dual-modal information interaction, we propose the SADMR conditioning mechanism. This mechanism enables intermediate features, incorporating prior knowledge from both modalities, to adaptively and iteratively attend to each modality's features during the denoising process. The visible images $ I_v $ and the infrared images $ I_i $ are first embedded by the pre-trained encoder $\varepsilon$ to get the representation $ {L_v} = \varepsilon (I_v) $ and $ {L_i} = \varepsilon (I_i) $. Then $ {L_v} $ and $ {L_i} $ are concatenated with the original noise $ z_T $ and, on the other branch, encoded to get the multi-scale features $ \{{F_v^l}\}_{l=1}^L $ and $ \{{F_i^l}\}_{l=1}^L $. The scale of $ \{{F_v^l}\}_{l=1}^L $ and $ \{{F_i^l}\}_{l=1}^L $ matches that of the corresponding multi-scale intermediate features $ \{{F_e^l}\}_{l=1}^L $ and $ \{{F_d^l}\}_{l=1}^L $ in the Unet-based denoising network, where $ {F_e^l} $ and $ {F_d^l} $ are respectively the features in the encoder and decoder modules. Before each downsampling and upsampling layer in the Unet-based denoising network, $ {F_v^l} $ and $ {F_i^l} $ with corresponding scale to intermediate features are parallelly injected. For better comprehension of the image features, rather than simple adding or concatenating operation, we leverage the deformable attention mechanism \cite{zhu2020deformable} to design a sparse cross-attention module (SCAM) for processing the injected features. The diagram of the sparse cross-attention module is shown in the supplementary material.

In SCAM, each element of intermediate features $ {F_e^l} $ and $ {F_d^l} $ selectively attends to certain elements surrounding its reference location in $ {F_v^l} $ and $ {F_i^l} $. Taking an arbitrary intermediate feature $ {F_e^l} $ and its corresponding visible image feature $ {{F_v^l}} $ as an example, these features are first encoded to generate multi-head queries and values $ \{ {Q^m}\} _{m = 1}^M,\{ {V^m}\} _{m = 1}^M $:
\begin{equation}
  \{ {Q^m}\} _{m = 1}^M = Q = {F_e^l}{W^Q},\{ {V^m}\} _{m = 1}^M = V = {F_v^l}{W^V},
  \label{eq:QV}
\end{equation}
where $ {W^Q} ,{W^V} $ are linear matrices. $ m $ indexes the attention head, and $ M $ is the number of attention heads.

Subsequently, the queries and values engage in deformable attention \cite{zhu2020deformable}. Arbitrary element $ i $ of  $ Q $ is used to characterize the deformable attention:

\begin{equation}
	\begin{split}
        {\rm{DeformAttn}}\left( {{Q_i},V} \right) = \sum\limits_{m = 1}^M {W_m}\Big[\sum\limits_{k = 1}^K {{\cal A}_k}{(Q_i^m)}\cdot
        \\{V^m}\left({r_i} + {{\cal O}_k}(Q_i^m)\right)\Big],
	\end{split}
 \label{eq:SCA}
\end{equation}

where $ k $ indexes the sampled element of the value map, $ K $ denotes the total sampled number, and $ {r_i} $ represents the 2D reference location of element $ i $ within $ Q $. $ W_m $ is a linear projection operator. $ {{\cal A}_k}( \cdot ) $ and $ {{\cal O}_k}( \cdot ) $ are linear blocks which are respectively utilized to calculate the weight and sampling offset of $ k $-th sampling point. Eventually, all the elements of the intermediate features will attend to their surrounding points in both infrared and visible images. 

\subsection{Information synthesis pretext task}
\label{sec:pretext}
Due to the difficulty of obtaining supervision to guide both information enhancement and complementation in the VIIS task, we design the ISPT to guide the model training process. ISPT leverages a data augmentation strategy involving a series of image processing methods to generate pseudo-low-light images from high-quality day-time visible images. The pseudo-low-light images are severely degraded, exhibiting reduced overall visibility and the obliteration of specific regions. In the training process, the model takes the pseudo-low-light images and corresponding infrared images from the dataset as inputs to recover the original high-quality visible images. The model is directed to enhance the overall quality of visible images while complementing the obliterated regions by leveraging information from the infrared modality and colorizing it with nearly authentic color. 

In the data augmentation process, original visible images undergo two image transformation methods along with the addition of certain noise. Specifically, the augmentation pipeline contains the following steps:

(1) Utilizing intense gamma transform to darken daytime images and obliterate certain regions:
\begin{equation}
  {I_g} = I_o^\gamma,
  \label{eq:Gamma}
\end{equation}
where $ I_o $ is the original clear visible image, and $ \gamma $ is the gamma parameter. When $ \gamma  > 1 $, the transformation is biased towards lower (darker) grayscale pixel intensities. In this process, regions with relatively low original luminance are naturally occluded, aligning closely with the inductive bias of practically captured images.

(2) Implementing contrast adjustment to reduce overall visibility and emulate the color and contrast feature in realistic low-light situations:
\begin{equation}
  {I_c} = \alpha {I_g},
  \label{eq:Contrast}
\end{equation}
where $ \alpha $ is the contrast factor which induces a reduction in contrast when $ \alpha  < 1 $. This operation is designed to imitate the low contrast and color fading in low-light environments while motivating the model to address these problems.

(3) Gaussian and Poisson noises are introduced to further corrupt visible image information. These simulate the electronic and shot noise present in real low-light images while prompting the model to resort to infrared images.

\begin{equation}
  {I_v} = {I_c} + {\cal N}(0, \sigma) + {\cal P}(\lambda),
  \label{eq:Noise}
\end{equation}

\section{Experiments}
In this section, we present detailed experimental settings, including descriptions of training and test datasets, along with implementation details. Then we quantitatively and qualitatively evaluate our method against state-of-the-art methods across relevant domains. Additionally, we construct new baselines capable of achieving both multi-modal information enhancement and fusion, and compare our model with these new baselines. Furthermore, we conduct extensive ablation experiments.

\begin{figure*}
\centering
\includegraphics[width=0.95\linewidth]{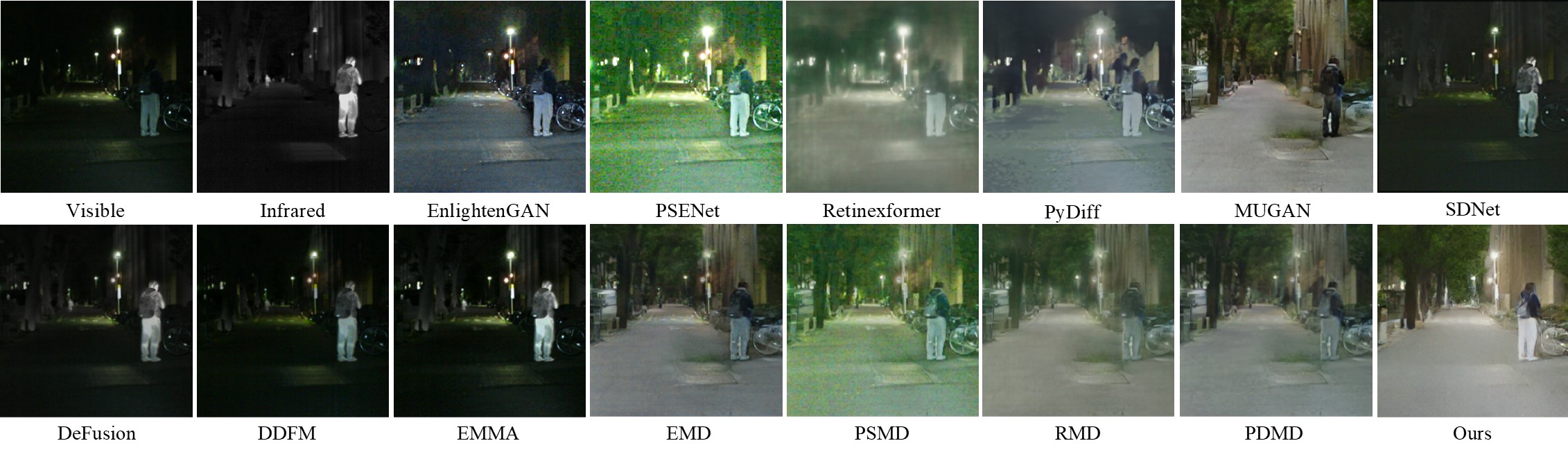}
   \caption{Qualitative comparison of '01012N' from MSRS dataset.}
\label{fig:MSRS1}
\end{figure*}

\begin{figure*}
\centering
\includegraphics[width=0.95\linewidth]{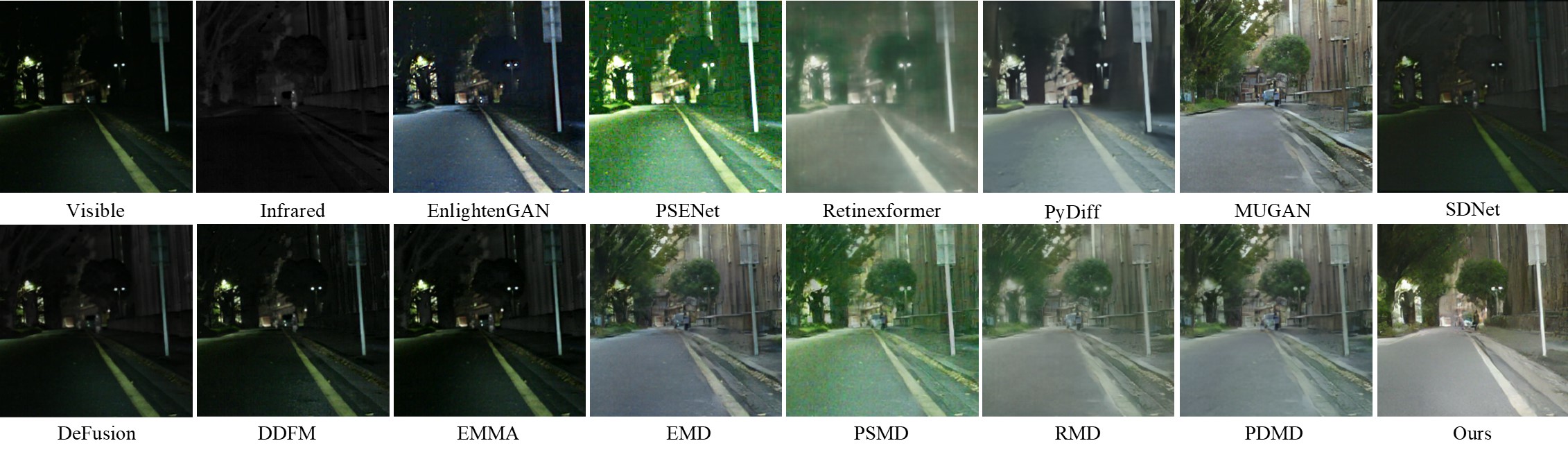}
   \caption{Qualitative comparison of '01198N' from MSRS dataset.}
\label{fig:MSRS2}
\end{figure*}

\begin{figure*}[h]
\centering
\includegraphics[width=0.95\linewidth]{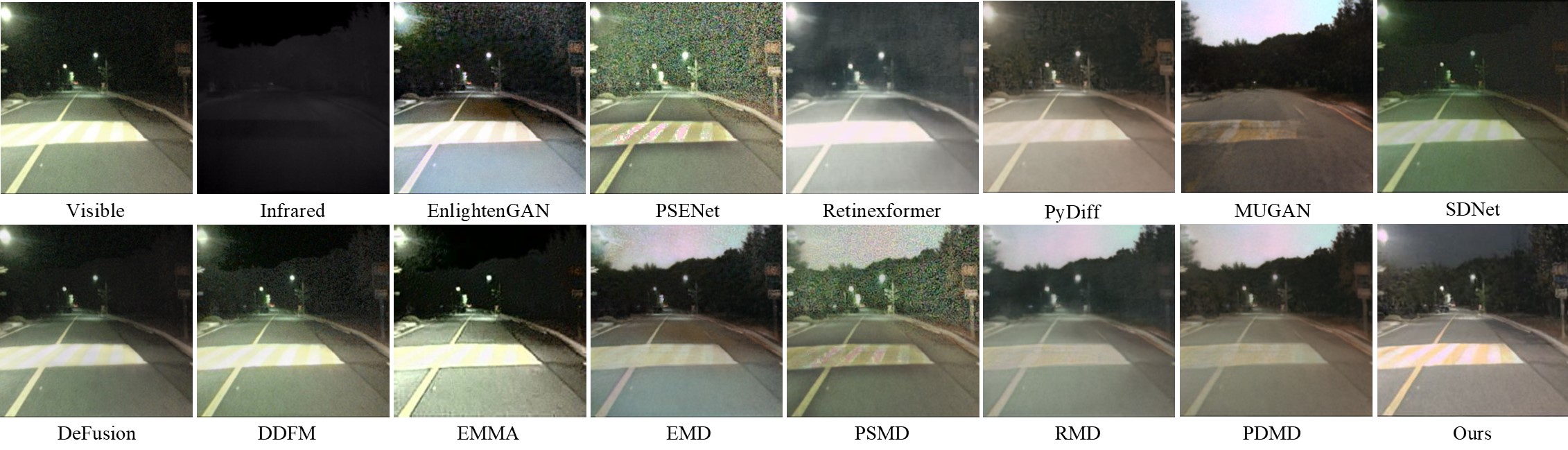}
   \caption{Qualitative comparison of 'set09\_v000\_00330' from KAIST-MS dataset.}
\label{fig:KAIST1}
\end{figure*}

\begin{figure*}[h]
\centering
\includegraphics[width=0.95\linewidth]{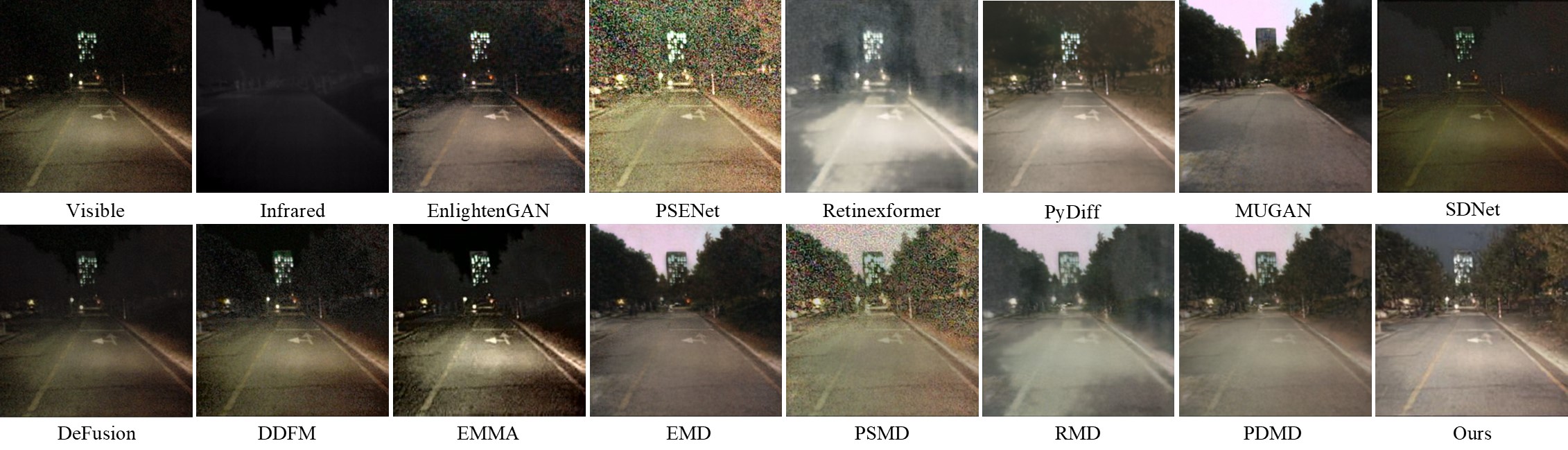}
   \caption{Qualitative comparison of 'set09\_v000\_00805' from KAIST-MS dataset.}
\label{fig:KAIST2}
\end{figure*}

\subsection{Datasets and experimental settings }
\label{sec:setting}
\noindent \textbf{Datasets.} We evaluate our model on two public datasets, MSRS \cite{tang2022piafusion}, KAIST-MS \cite{hwang2015multispectral}. These datasets comprise aligned visible and infrared image pairs, where the daytime and nighttime images are accurately separated.  Our model is trained using the daytime training sets of MSRS (536 pairs) and KAIST-MS (33,399 pairs) respectively. 

For testing our model and aligning with the task requirements, we specifically select image pairs where the visible images are severely degraded by realistic low-light conditions. As a result, we obtain 62 image pairs from the testing set of MSRS and 68 image pairs from the testing set of KAIST-MS.

\noindent \textbf{Baselines.}
Since VIIS is a novel task, to demonstrate the superiority of VIIS in enhancing severe low-light images, diverse methods in relevant fields are first compared with our model. Specifically, we compare our model against four low-light image enhancement methods: the unpaired method EnlightenGAN \cite{jiang2021enlightengan}, the supervised method Retinexformer \cite{Cai_2023_ICCV} PyDiff \cite{ijcai2023p199} and the unsupervised method PSENet \cite{nguyen2023psenet}, one infrared image colorization method MUGAN \cite{liao2022mugan} and four infrared and visible image fusion methods consisting of SDNet \cite{zhang2021sdnet} DeFusion \cite{liang2022fusion} DDFM \cite{Zhao_2023_ICCV} and EMMA \cite{zhao2024equivariant}. Since there is no ground truth in MSRS \cite{tang2022piafusion} and KAIST-MS \cite{hwang2015multispectral} dataset, for the supervised method Retinexformer \cite{Cai_2023_ICCV} and PyDiff \cite{ijcai2023p199}, we utilize the data augmentation method of the ISPT to construct paired training images, where the parameter configuration adheres to that of our training process. 

In addition, to intuitively evaluate the performance of our model, we construct new baselines that can achieve both dual-modal information enhancement and fusion. The new baselines utilize low-light image enhancement and infrared image colorization methods to enhance and colorize visible and infrared images, respectively. Subsequently, these baselines employ image fusion techniques to merge the outputs of the low-light image enhancement and infrared image colorization methods. For the fusion model, we select the versatile self-supervise-based image fusion model, DeFusion \cite{liang2022fusion}. For the low-light image enhancement and infrared image colorization models, we involve all the aforementioned models EnlightenGAN \cite{jiang2021enlightengan}, Retinexformer \cite{Cai_2023_ICCV}, PSENet \cite{nguyen2023psenet}, PyDiff \cite{ijcai2023p199}, and MUGAN \cite{liao2022mugan}. We design four new baselines as shown in \cref{tab:baseline}

\begin{table}[]
  \centering
    {\footnotesize{
\begin{tabular}{llll}
\toprule
Baseline & Enhancement &  Colorization & Fusion \\
\midrule
EMD & EnGAN \cite{jiang2021enlightengan} & \multirow{4}{*}{MUGAN \cite{liao2022mugan}} & \multirow{4}{*}{DeFusion \cite{liang2022fusion}} \\
PSMD & PSENet \cite{nguyen2023psenet} &  &  \\
RMD & Retinexformer \cite{Cai_2023_ICCV} &  &  \\
PDMD & PyDiff \cite{ijcai2023p199} &  &  \\
\bottomrule
\end{tabular}
}}
\caption{Our newly designed baselines.}
\label{tab:baseline}
\end{table}
\noindent \textbf{Implementation details.} Our experiments are conducted on a single NVIDIA RTX3090 GPU. The parameters of the network are optimized by the Adam optimizer, with the learning rate set to $ 1.6 \times {10^{-5}} $ and a batch size of 8. We crop and resize the input image pairs to $ 256 \times 256 $. The downsampling factor $ f $ is set as 4 and the length of the denoising step $ T $ is set as 200. In the training process, we configure the gamma transform parameter $ \gamma $ within the interval [3, 10], the contrast factor $ \alpha $ in the range of [0.1,1], the standard deviation $ \sigma $ of Gaussian noise within the range of [0,10], and the mean parameter $ \lambda $ of the Poisson noise within range of [0,20].

\subsection{Model comparison.}
\noindent \textbf{Qualitative comparison.}  \cref{fig:MSRS1,fig:MSRS2} show the visual comparison results with different baselines on the MSRS dataset. \cref{fig:KAIST1,fig:KAIST2} show the visual comparison results with different baselines on the KAIST-MS dataset. It can be seen that for the low-light enhancement methods, all of the three methods are unable to restore these severely corrupted image regions and produce artifacts. In addition, PSENet \cite{nguyen2023psenet} results in noticeable color deviations and noise, while Retinexformer \cite{Cai_2023_ICCV} introduces blurring in certain regions. The infrared image colorization method MUGAN \cite{liao2022mugan} produces unnatural artifacts, while failing to restore authentic color. In addition, limited by the sketchy infrared images, it loses some vital texture details \eg road markings and light. In the case of infrared and visible image fusion methods, the fused images still suffer from color distortion and low luminance while the information from the infrared images remains grayscale and sketchy. Complemented by the colorized infrared images, the newly designed baselines outperform singular modality low-light image enhancement methods in restoring severely corrupted image regions. However, limited by the independent processes of enhancement, colorization, and fusion, these new baselines fail to address issues inherent in the sub-models (\ie, low-light image enhancement and infrared image colorization models). As a result, problems such as unauthentic color and texture information, noise, and artifacts persist. In contrast, our model effectively restores the 
severely corrupted image regions, maintains authentic image color, and enhances the overall perceptual quality.

\noindent \textbf{Quantitative comparison.}
We adopt standard deviation (SD) $\uparrow$, entropy (EN) $\uparrow$, natural image quality evaluator (NIQE) $\downarrow$ \cite{mittal2012making}, blind/referenceless image spatial quality evaluator (BRISQUE) $\downarrow$ \cite{mittal2012no} as metrics. EN measures the amount of information, SD assesses the contrast, while NIQE and BRISQUE evaluate the comprehensive perceptual quality. As shown in \cref{tab:msrs,tab:kaist}, our model achieves the best performance, except for the SD metric in the KAIST-MS dataset. In that specific case, our model attains the second-best result, with MUGAN \cite{liao2022mugan} achieving the best result. This is mainly because MUGAN transforms the infrared images to the visible daytime domain, achieving high contrast.

\begin{table}[hbt]
  \centering
    {\small{
\begin{tabular}{lllll}
\toprule
Method & SD & EN & NIQE & BRISQUE  \\
\midrule
EnGAN \cite{jiang2021enlightengan} & 40.76 & 6.81 & 4.29 & \underline{20.33}\\
PSENet \cite{nguyen2023psenet} & \underline{45.76} & 6.98 & \underline{3.61} & 24.72\\
Reformer \cite{Cai_2023_ICCV} & 37.74 & 6.76 & 6.75 & 37.93\\
PyDiff \cite{ijcai2023p199} & 38.77 & 7.06 & 5.64 & 37.16 \\
MUGAN \cite{liao2022mugan} & 44.69 & 6.96 & 4.78 & 29.59\\
SDNet \cite{zhang2021sdnet} &  17.41 & 5.08 & 6.01 & 30.24\\
DeFusion \cite{liang2022fusion} & 22.99 & 5.38 & 5.89 & 28.02\\
DDFM \cite{Zhao_2023_ICCV} & 22.21 & 5.31 & 4.77 & 20.69\\
EMMA \cite{zhao2024equivariant} & 29.25 & 5.75 & 5.38 & 21.84\\
EMD \cite{jiang2021enlightengan,liang2022fusion,liao2022mugan} & 38.19 & 6.88 & 4.12 & 25.41\\
PSMD \cite{nguyen2023psenet,liang2022fusion,liao2022mugan}& 37.74 & \underline{7.14} & 3.91 & 22.76\\
RMD \cite{Cai_2023_ICCV,liang2022fusion,liao2022mugan}& 34.28 & 6.88 & 4.55 & 27.13\\
PDMD \cite{ijcai2023p199,liang2022fusion,liao2022mugan}& 35.57 & 7.10 & 4.33 & 26.50\\
Ours & \textbf{48.81} & \textbf{7.15} & \textbf{3.57} & \textbf{14.44}\\
\bottomrule
\end{tabular}
}}
\caption{Quantitative results on the MSRS \cite{tang2022piafusion} dataset. The best, second-best results are marked with \textbf{bold}, and \underline{underlined}.}
\label{tab:msrs}
\vspace{-1.0em}
\end{table}

\begin{table}
  \centering
    {\small{
\begin{tabular}{lllll}
\toprule
Method & SD & EN & NIQE & BRISQUE\\
\midrule
EnGAN \cite{jiang2021enlightengan} & 38.82 & 6.91 & 4.20 & 26.51\\
PSENet \cite{nguyen2023psenet} & 37.30 & 6.96 & \underline{3.82} & 30.05\\
Reformer \cite{Cai_2023_ICCV} & 31.17 & 6.82 & 4.46 & 31.07\\
PyDiff \cite{ijcai2023p199} & 37.10 & 6.94 & 4.29 & 28.37 \\
MUGAN \cite{liao2022mugan} & \textbf{50.59} & 6.71 & 6.14 & 28.61\\
SDNet \cite{zhang2021sdnet} &  20.16 & 5.51 & 5.58 & 33.49\\
DeFusion \cite{liang2022fusion} & 20.93 & 6.60 & 5.34 & 28.48\\
DDFM \cite{Zhao_2023_ICCV} & 22.95 & 6.65 & 4.39 & 24.40\\
EMMA \cite{zhao2024equivariant} & 25.25 & 7.01 & 5.12 & 24.72\\
EMD \cite{jiang2021enlightengan,liang2022fusion,liao2022mugan}& 44.77 & 7.05 & 3.95 & 27.12\\
PSMD \cite{nguyen2023psenet,liang2022fusion,liao2022mugan}& 44.17 & \underline{7.06} & 4.24 & \underline{24.24}\\
RMD \cite{Cai_2023_ICCV,liang2022fusion,liao2022mugan}& 40.91 & 6.93 & 5.09 & 30.37\\
PDMD \cite{ijcai2023p199,liang2022fusion,liao2022mugan}& 41.98 & 7.05 & 4.14 & 26.38\\
Ours & \underline{46.30} & \textbf{7.11} & \textbf{3.67} & \textbf{16.08}\\
\bottomrule
\end{tabular}
}}
\caption{Quantitative results on the KAIST-MS \cite{hwang2015multispectral} dataset. The best, second-best results are marked with \textbf{bold}, and \underline{underlined}.}
\label{tab:kaist}
\end{table}

\subsection{Ablation studies}
For ablation experiments, we train and test our model with different configurations on MSRS dataset. The results are shown in \cref{tab:ablation}.

\noindent \textbf{Sparse cross attention module.} We explore the impact of the sparse cross-attention module by substituting it with the addition, concatenation, and global cross-attention operations. In this configuration, the three conditioning images are also initially encoded and concatenated with the original noise as depicted in \cref{sec:conditioning}. In the UNet-based denoising network, the encoded features are incorporated via addition, concatenation, and global cross-attention operations. As shown in \cref{tab:ablation}, our model performs better than these three methods with other configurations demonstrating the effectiveness of the SCAM in processing the injected features.

\noindent \textbf{SADMR conditioning mechanism.} To assess the impact of the SADMR conditioning mechanism, we remove it, resulting in the conditioning images solely being concatenated with the input noise in the Unet-based denoising network. As shown in \cref{tab:ablation} our original model achieves better performance, demonstrating the capacity of SADMR conditioning mechanism for effective information interaction. However, the model in this configuration achieves better performance than the models in the configurations SCAM $ \rightarrow $ Addition and SCAM $ \rightarrow $ Concatenation. We infer that this phenomenon is mainly because these simple injection operations introduce additional noise while injecting features.

\noindent \textbf{Visible and infrared images.} To evaluate the contribution of visible and infrared images in the multi-model information synthesis process, we respectively remove each modality, leaving the model with the same input of either visible or infrared images. As shown in \cref{tab:ablation},  the exclusion of either modality results in a decline in all quantitative performance metrics.

\begin{table}

  \centering
    {\footnotesize{
\begin{tabular}{cllll}
\toprule
    Configurations & SD & EN & NIQE & BRISQUE\\
    \midrule
    SCAM $ \rightarrow $ Addition & 46.22 & 6.96 & 3.89 & 16.88 \\
    SCAM $ \rightarrow $ Concatenation & 46.45 & 6.98 & 3.84 & 16.91 \\
    SCAM $ \rightarrow $ Global attention & 47.03 & 6.98 & 3.73 & 16.04 \\
    w/o SADMR & 46.64 & 7.01 & 3.78 & 16.59\\
    w/o Visible & 44.97 & 7.05 & 4.47 & 30.43\\
    w/o infrared & 41.14 & 6.77 & 5.87 & 29.10\\
    
    Original & \textbf{48.81} & \textbf{7.15} & \textbf{3.57} &  \textbf{14.44}\\
   
   \bottomrule
\end{tabular}
}}
\caption{Ablation experiment results on MSRS dataset. The best results are marked with \textbf{bold}.}
\label{tab:ablation}
\end{table}

\section{Limitation and broader impact}
\begin{figure}
\centering
\includegraphics[width=0.7\linewidth]{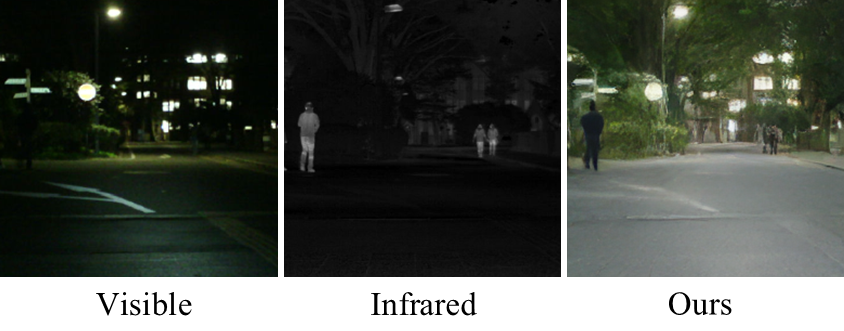}
   \caption{In this instance, although our method enhances the overall perceptual quality and reveals the targets hidden in darkness, the ground indicator unique to the visible image is weakened in the output.}
\label{fig:Limitation}
\end{figure}

\noindent \textbf{Limitation.}
To tackle the problem of lacking supervision in VIIS, we design the ISPT. However, compared to the pretext task in more established fields, ours is relatively simple and far from perfect. We have observed that although this pretext guides the model to leverage infrared information to complement corrupted visible counterparts, it may weaken information that is unique to visible images (\eg, texture or color information). An instance of this is illustrated in \cref{fig:Limitation}. However, our pretext task provides a new approach to lead comprehensive information enhancement and fusion. We hope it will inspire more work in the related communities.

\noindent \textbf{Broader impact.} 
Our model introduces a novel paradigm that unifies information enhancement and fusion for visible and infrared images in an end-to-end model. When presented with a visible image severely degraded by low light, a scenario very challenging for single visible modality-based enhancement methods, along with the corresponding infrared image, our model can effectively reveal situational information and achieve high perceptual quality. Consequently, it has promising applications in low-light imaging fields, particularly those that rely heavily on thermal infrared information, such as nighttime surveillance systems \cite{ibrahim2016comprehensive} and wildlife photography \cite{cilulko2013infrared, burke2019optimizing}.

\section{Conclusion}
In this paper, we propose a novel task, VIIS. It achieves fusing and enhancing information from visible and infrared images simultaneously. To address the challenge of lacking ground truth in VIIS, we design the ISPT based on an image augmentation strategy. Additionally, we introduce the SADMR conditioning mechanism for the diffusion model to enable effective information interaction between the two modalities. Our extensive experiments demonstrate that our method outperforms existing state-of-the-art methods in related fields (\ie, low-light image enhancement, infrared image colorization, and infrared and visible image fusion) for enhancing realistic severe low-light images. Furthermore, for a more comprehensive and intuitive evaluation, we compare our method with newly designed baselines which integrate these relevant methods and can enhance and fuse information from the two modalities. The results show that our model also qualitatively and quantitatively outperforms the new baselines.
\newpage
\twocolumn[
\begin{center}
    {\LARGE Supplementary Materials} 
\end{center}
]

\setcounter{section}{6}
\setcounter{figure}{7}
\setcounter{table}{4}
\section{Ablation studies for ISPT}
\begin{figure}[b]
\centering
\includegraphics[width=1\linewidth, trim={0mm 0mm 0mm 0.5mm}, clip]{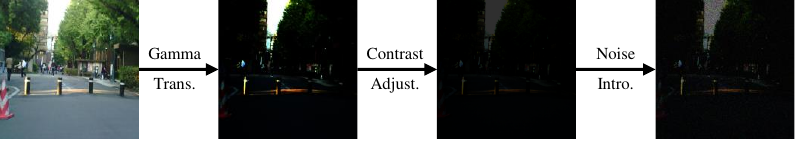}
   \caption{The data augmentation process of ISPT.}
\label{fig:ISPT}
\end{figure}

In this section, we conduct ablation experiments to assess the individual contributions of components in ISPT. \cref{fig:ISPT} illustrates the data augmentation process of ISPT, where each component is systematically ablated to verify its effectiveness.

Specifically, we first reduce the influence of the gamma transform and contrast adjustment operations separately. For the gamma transform, the parameter $ \gamma $ is constrained within the range [1, 5], and the contrast adjustment factor $ \alpha $ is limited to [0.55, 1]. In addition, we respectively conduct experiments by completely removing the gamma transform and contrast adjustment to further evaluate their roles. Finally, the noise introduction process is respectively replaced with impulse noise with the noise density being in the range of [0, 0.2] and directly removed to evaluate its contribution.

\begin{figure}[b]
\centering
\includegraphics[width=1\linewidth]{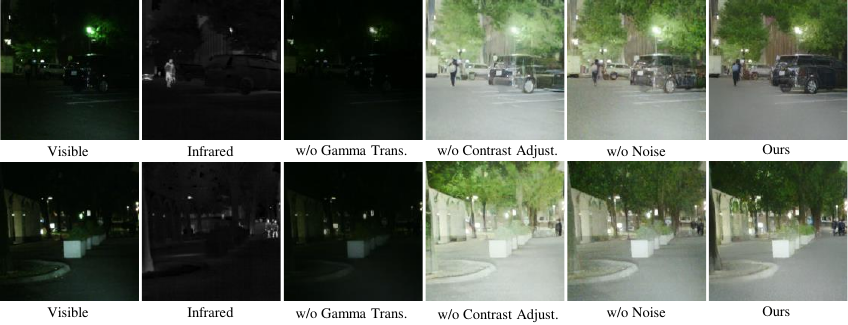}
   \caption{Qualitative results on discussing components of ISPT.}
\label{fig:ISPT_ab}
\end{figure}

As shown in \cref{tab:ablation}, our model outperforms alternative configurations of ISPT, demonstrating the effectiveness of its individual components. Additionally, \cref{fig:ISPT_ab} highlights key observations: the absence of the gamma transform significantly impairs the model’s ability to enhance image quality. When contrast adjustment is omitted, the generated images exhibit brightness and color distortions. Finally, the exclusion of noise introduction of ISPT reduces the model's ability to effectively address the inherent noise in the original images.

\begin{table}[b]

  \centering
    {\footnotesize{
\begin{tabular}{cllll}
\toprule
    Configurations & SD$\uparrow$ & EN$\uparrow$ & NIQE$\downarrow$ & BRISQUE$\downarrow$\\
    \midrule
    Reduced $ \gamma $ & 45.56 & 7.04 & 4.02 & 20.72 \\
    Reduced $ \alpha $ & 44.60 & 7.13 & 3.68 & 18.74 \\
    w/o Gamma Transform & 12.71 & 4.47 & 9.01 & 37.91 \\
    w/o Contrast Adjustment & 42.49 & 7.08 & 3.91 & 22.01 \\
    Impulse Noise & 46.02 & 6.94 & 3.67 & 17.51\\
    w/o Noise & 46.92 & \textbf{7.23} & 4.00 & 20.96\\
    
    Original & \textbf{48.81} & 7.15 & \textbf{3.57} &  \textbf{14.44}\\
   
   \bottomrule
\end{tabular}
}}
\caption{Ablation experiment results on MSRS dataset. The best results are marked with \textbf{bold}.}
\label{tab:ablation}
\end{table}

\section{User studies}
\begin{figure}[t]
\centering
\includegraphics[width=1\linewidth]{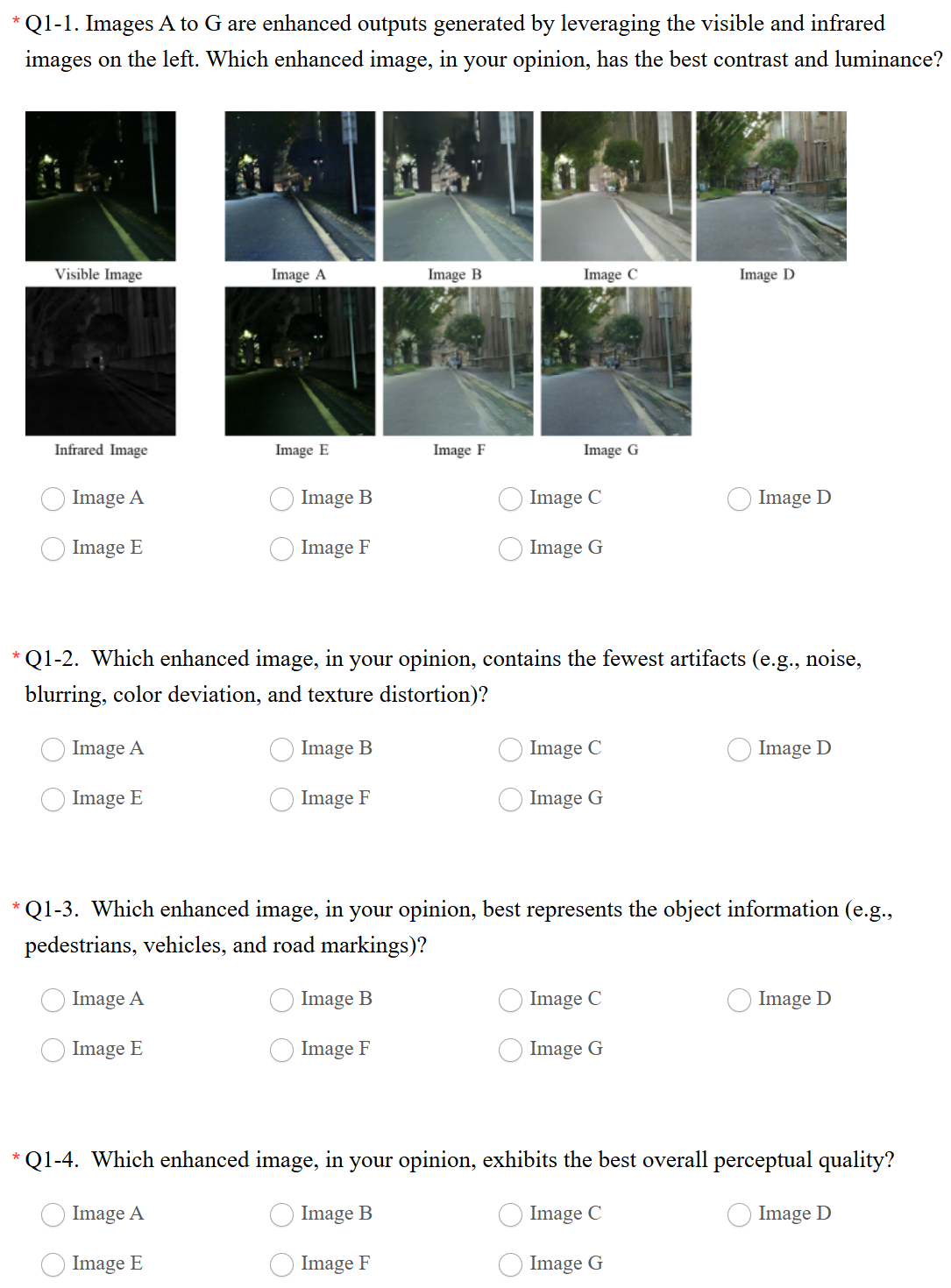}
   \caption{Screenshot of the questionary interface from user study for different models in different dimensions.}
\label{fig:qusetionary}
\end{figure}

We conduct user studies to evaluate the performance of our method. We select 20 images from the testing set of MSRS \cite{tang2022piafusion}, KAIST-MS \cite{hwang2015multispectral}. For each image, it is enhanced by seven methods (EnlightenGAN \cite{jiang2021enlightengan}, PyDiff \cite{ijcai2023p199}, EMMA \cite{zhao2024equivariant}, EMD \cite{jiang2021enlightengan,liang2022fusion,liao2022mugan}, PDMD \cite{ijcai2023p199,liang2022fusion,liao2022mugan}, and ours). Subsequently, 15 participants are invited to choose the best method regarding contrast and luminance, artifacts (\eg noise, blurring, color deviation, and texture distortion), object information (\eg pedestrians, vehicles, and road markings), and overall perceptual quality. All of the participants have good English proficiency and most of them have basic knowledge of computer vision. An example of the questionary interface for different models in different dimensions can be seen in \cref{fig:qusetionary}.

\begin{figure}[h]
  \centering
  \includegraphics[width=1\linewidth]{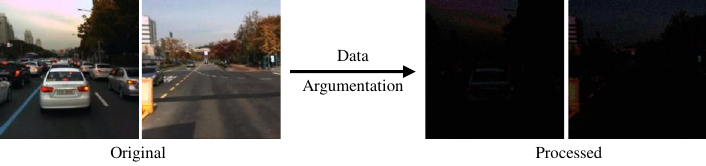}

   \caption{Some instances in the testing dataset where the high-quality original images undergo data augmentation to get the pseudo-low-light images.}
\label{fig:Argumentation}
\end{figure}

As shown in \cref{tab:user study}, except for the salience of object information, our model achieves much better results than the other six methods. Although VIF methods can generate images with thermal radiation foreground objects contrasting with the background, more participants think our model achieves better overall object salience.

\begin{table*}[h]
  \centering
     {\footnotesize{
\begin{tabular}{cccccccc}
\toprule
Dimension & EnGAN \cite{jiang2021enlightengan} & PyDiff \cite{ijcai2023p199} & MUGAN \cite{liao2022mugan} & EMMA \cite{zhao2024equivariant} & EMD & PDMD & Ours\\
\midrule
Contrast and Luminance & 5.33\% & 7.67\% & 11\% & 0.33\% & 7\% & 10.33\% & \textbf{58.33\%}\\
Artifacts & 4\% & 8.67\% & 3\% & 6\% & 5\% & 6.33\% & \textbf{67\%}\\
Object Information & 2.33\% & 3\% & 6.67\% & 31\% & 6\% & 7.67\% & \textbf{43.33\%}\\
Overall Quality & 7\% & 8.67\% & 5.67\% & 5.67\% & 8\% & 9.67\% & \textbf{55.33\%}\\
\bottomrule
  \end{tabular}
  }}
  \caption{The result for user study, where the values represent the percentage of votes obtained by each method in a certain dimension for all images and participants. The best result is marked with \textbf{bold}.}
  \label{tab:user study}
\end{table*}

\section{Analysis for information fidelity}
\label{sec:intro}
To evaluate the information fidelity between the model outputs and authentic information, in this part, we conduct experiments to test the consistency between output images and ground truth. Since the absence of ground truth in the existing aligned visible and infrared image datasets, we construct a synthetic dataset. Specifically, we conduct the fidelity experiment on KAIST-MS \cite{hwang2015multispectral} dataset and employ the image augmentation method of the information synthesis pretext task (ISPT) to degrade clear high-quality images in the daytime testing set. Using the degraded pseudo-low-light visible images and the corresponding infrared images, our model and baselines are employed to regenerate the original images. Subsequently, we calculate the Structural Similarity Index (SSIM) and Peak Signal-to-Noise Ratio (PSNR) metrics between the generated images and the original images.

\begin{table}[h]
  \centering
     {\footnotesize{
\begin{tabular}{llll}
\toprule
Baseline & Enhancement &  Colorization & Fusion \\
\midrule
EMD & EnGAN \cite{jiang2021enlightengan} & \multirow{4}{*}{MUGAN \cite{liao2022mugan}} & \multirow{4}{*}{Defusion \cite{liang2022fusion}} \\
UMD & Uformer \cite{9878729} &  &  \\
RMD & Reformer \cite{Cai_2023_ICCV} &  &  \\
PDMD & PyDiff \cite{ijcai2023p199} &  &  \\
\bottomrule
  \end{tabular}
  }}
  \caption{The newly designed baselines for information fidelity experiment.}
  \label{tab:baseline}
\end{table}

\subsection{Experimental settings }
\noindent \textbf{Baselines.} We compare our model against four low-light image enhancement methods EnlightenGAN \cite{jiang2021enlightengan}, Uformer \cite{9878729}, Retinexformer \cite{Cai_2023_ICCV} and PyDiff \cite{ijcai2023p199} and one infrared image colorization method MUGAN \cite{liao2022mugan}. In addition, we also utilize the versatile self-supervise-based image fusion model, Defusion \cite{liang2022fusion} to combine the low-light image enhancement and infrared image colorization methods to build up new baselines which are shown in \cref{tab:baseline}.

\begin{table*}[!h]
  \centering
     {\footnotesize{
\begin{tabular}{ccccccccccc}
\toprule
Metric & EnGAN \cite{jiang2021enlightengan} &  Uformer \cite{9878729} & Reformer \cite{Cai_2023_ICCV} & PyDiff \cite{ijcai2023p199} & MUGAN \cite{liao2022mugan} & EMD & UMD & RMD & PDMD & Ours\\
\midrule
SSIM & 0.52 & 0.59 & 0.59 & 0.62 & 0.54 & 0.59 & 0.61 & 0.62 & 0.62 & \textbf{0.69}\\
PSNR & 14.35 & 15.71 & 18.87 & 17.44 & 16.29 & 16.56 & 17.53 & 18.74 & 18.44 & \textbf{21.88}\\
\bottomrule
  \end{tabular}
  }}
  \caption{Quantitative results for fidelity experiment. The best result is marked with \textbf{bold}.}
  \label{tab:quantitive}
\end{table*}

\begin{figure*}[!h]
  \centering
  \includegraphics[width=0.95\linewidth]{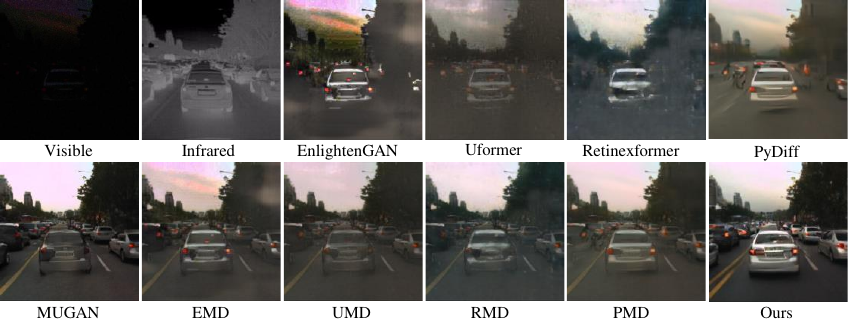}

   \caption{Qualitative comparisons for fidelity experiment. The ground truth is shown on the left of \cref{fig:Argumentation}.}
\label{fig:Supplementary}
\end{figure*}

\noindent \textbf{Implementation details.} The training dataset and parameter configuration of the data argumentation of our model and the four low-light image enhancement methods are the same as those in Sec. 4.1 of the main paper. For the testing dataset, as the KAIST-MS dataset is derived from videos and displays frame similarities, we select one image pair from every 400 frames to construct the testing dataset. Additionally, the parameters of the image augmentation for the testing dataset are set as follows: the gamma parameter $ \gamma $ is configured at 6, the contrast factor $ \alpha $ is set to 0.5, and the noise parameters are set as $ \lambda  = 10, \sigma  = 5 $. Some instances in the testing dataset are shown in \cref{fig:Argumentation}.

\subsection{Model comparisons.}

As shown in \cref{fig:Supplementary}, the low-light image enhancement methods fail to restore these severely corrupted image regions which lack essential information and introduce artifacts and blurring. The infrared image colorization method MUGAN \cite{liao2022mugan} is unable to generate authentic color and texture information. Although the newly designed baselines realize information complementation between the above two sorts of methods, they can't avoid the intrinsic problems in them (\eg, unauthentic color and artifacts). In contrast, our method restores the processed image to the original with high fidelity. Additionally, according to \cref{tab:quantitive}, our model outperforms all of the aforementioned methods with significant margins (+3 and +0.07 in PSNR and SSIM metrics, respectively).

\section{Diagram of sparse cross-attention module}
\begin{figure}[t]
  \centering
  \includegraphics[width=1\linewidth]{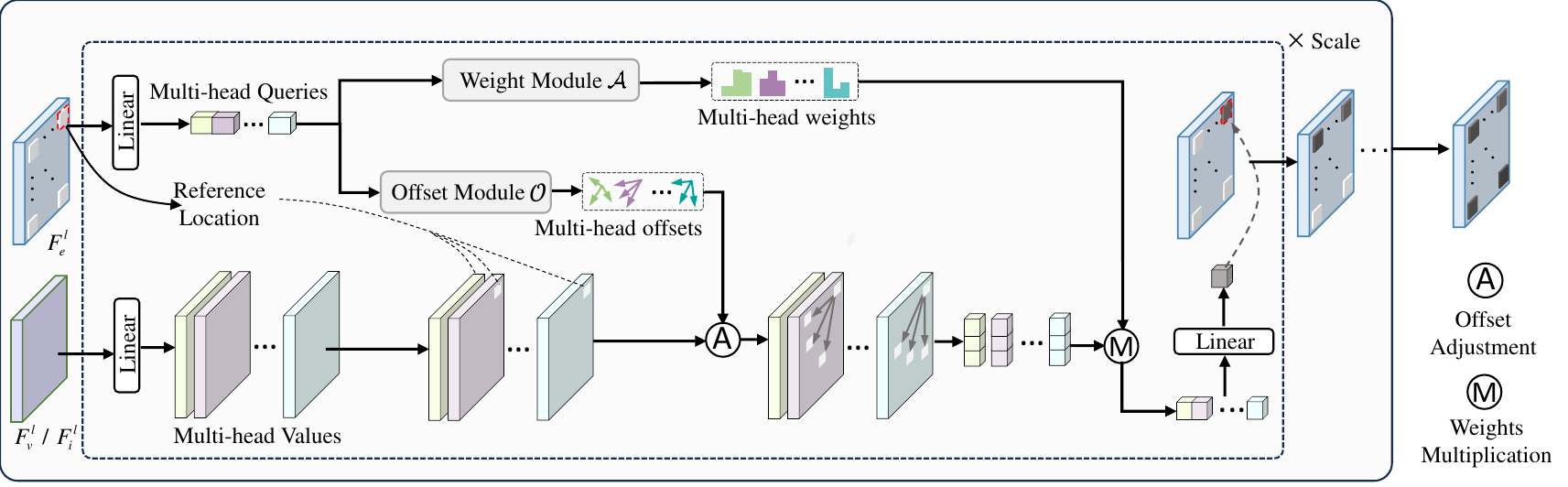}
   \caption{In SCAM, the intermediate feature $ {F_e^l} $ from the Unet-based denoising network and visible/infrared features $ {F_v^l}, {F_i^l}$ are initially embedded to get muti-head queries and values. Subsequently, each query element attends to elements surrounding its reference location in values, utilizing offset module ${\cal O}$ and weight module ${\cal A}$, both of which are linear blocks. Finally, the muti-head outputs undergo a linear projection operation to update the element of $ {F_e^l} $.}
\label{fig:SCAM}
\end{figure}

In this section, as shown in \cref{fig:SCAM}, we illustrate the sparse cross-attention module with an intermediate feature $ {F_e^l} $ of the encoder of the Unet-based denoising network.

\FloatBarrier

{\small
\bibliographystyle{ieee_fullname}
\bibliography{main}
}

\end{document}